\documentclass[sigconf]{acmart}
\AtBeginDocument{%
  \providecommand\BibTeX{{%
    \normalfont B\kern-0.5em{\scshape i\kern-0.25em b}\kern-0.8em\TeX}}}

\setcopyright{acmcopyright}
\copyrightyear{2018}
\acmYear{2022}
\acmDOI{XXXXXXX.XXXXXXX}

\usepackage{tabto}
\def\theLetterSpace{0.5pt}
\newcommand\spaceout[2][\theLetterSpace]{%
  \def\LocalLetterSpace{#1}\expandafter\spaceouthelpA#2 \relax\relax}
\def\spaceouthelpA#1 #2\relax{%
  \spaceouthelpB#1\relax\relax%
  \ifx\relax#2\else\ \kern\LocalLetterSpace\spaceouthelpA#2\relax\fi
}
\def\spaceouthelpB#1#2\relax{%
  #1%
  \ifx\relax#2\else
    \kern\LocalLetterSpace\spaceouthelpB#2\relax%
  \fi
}
\usepackage{microtype}
\usepackage{graphicx}
\usepackage{subfigure}
\usepackage{booktabs} 
\usepackage{hyperref}
\usepackage{url}
\usepackage{lipsum}
\usepackage{graphicx}
\usepackage{cleveref} 
\usepackage{listings}
\usepackage{xcolor}
\usepackage{enumitem}
\usepackage{booktabs}
\usepackage{xcolor}
\usepackage{colortbl}
\usepackage{pifont}
\usepackage[most]{tcolorbox}
\usepackage{makecell}
\usepackage[dvipsnames]{xcolor}
\newcommand{\cmark}{\textcolor{ForestGreen}{\ding{51}}} 
\newcommand{\xmark}{\textcolor{red}{\ding{55}}}         
\definecolor{headerred}{RGB}{220,0,0}

\definecolor{airforceblue}{rgb}{0.36, 0.54, 0.66}
\definecolor{burntumber}{rgb}{0.54, 0.2, 0.14}

\begin{document}

\title{Rethinking Failure Attribution in Multi-Agent Systems: A Multi-Perspective Benchmark and Evaluation}

\author{Yeonjun In}
\authornote{Work done during internship at Adobe Research.}
\email{yeonjun.in@kaist.ac.kr}
\affiliation{%
  \institution{KAIST}
  \country{Daejeon, Republic of Korea}}
  
\author{Md Mehrab Tanjim}
\authornote{Co-corresponding authors}
\email{tanjim@adobe.com}
\affiliation{\institution{Adobe Research}
\country{San Jose, CA, USA}}

\author{Jayakumar Subramanian}
\email{jasubram@adobe.com}
\affiliation{\institution{Adobe Research}
\country{San Jose, CA, USA}}

\author{Sungchul Kim}
\email{sukim@adobe.com}
\affiliation{\institution{Adobe Research}
\country{San Jose, CA, USA}}

\author{Uttaran Bhattacharya}
\email{ubhattac@adobe.com}
\affiliation{\institution{Adobe Research}
\country{San Jose, CA, USA}}

\author{Wonjoong Kim}
\email{wjkim@kaist.ac.kr}
\affiliation{\institution{KAIST}
\country{Daejeon, Republic of Korea}}

\author{Sangwu Park}
\email{sangwu.park@kaist.ac.kr}
\affiliation{\institution{KAIST}
\country{Daejeon, Republic of Korea}}

\author{Somdeb Sarkhel}
\email{sarkhel@adobe.com}
\affiliation{\institution{Adobe Research}
\country{San Jose, CA, USA}}

\author{Chanyoung Park}
\authornotemark[2]
\email{cy.park@kaist.ac.kr}
\affiliation{\institution{KAIST}
\country{Daejeon, Republic of Korea}}

\renewcommand{\shortauthors}{}
\newcommand{\proposed}{\textsc{MP-Bench}}
\newcommand{\yj}[1]{\textcolor{red}{~YJ:~#1}}
\begin{abstract}
Failure attribution is essential for diagnosing and improving multi-agent systems (MAS), yet existing benchmarks and methods largely assume a single deterministic root cause for each failure. In practice, MAS failures often admit multiple plausible attributions due to complex inter-agent dependencies and ambiguous execution trajectories. We revisit MAS failure attribution from a multi-perspective standpoint and propose \textit{multi-perspective failure attribution}, a practical paradigm that explicitly accounts for attribution ambiguity. To support this setting, we introduce \proposed, the first benchmark designed for multi-perspective failure attribution in MAS, along with a new evaluation protocol tailored to this paradigm. Through extensive experiments, we find that prior conclusions suggesting LLMs struggle with failure attribution are largely driven by limitations in existing benchmark designs. Our results highlight the necessity of multi-perspective benchmarks and evaluation protocols for realistic and reliable MAS debugging. Source code and dataset is presented in \href{https://github.com/yeonjun-in/MP-Bench}{\textcolor{magenta}{https://github.com/yeonjun-in/MP-Bench}}.
\end{abstract}




\maketitle

\section{Introduction}
Multi-agent systems (MAS) enable multiple LLM-based agents to collaborate through complex interactions to solve tasks across diverse domains \citep{wu2024autogen, fourney2024magentic, hong2023metagpt, ghafarollahi2025sciagents, zhang2024aflow}. Despite their increasing adoption, MAS continue to exhibit frequent execution failures, making it essential to understand where and why these failures occur \citep{cemri2025multi}. To this end, recent work has focused on \textit{failure attribution}, which aims to identify the failure-inducing steps from execution traces; however, existing approaches typically rely on manual, step-by-step inspection of agent behaviors, making them labor-intensive and time-consuming. To mitigate this burden, recent studies have explored automating failure attribution in MAS \cite{zhangagent, zhang2025agentracer, zhang2025graphtracer, kong2025aegis, ma2025automatic, west2025abduct}. A pioneering effort is Who\&When \cite{zhangagent}, the first benchmark for failure attribution in MAS, which labels each execution with a single, deterministic failure-inducing step for the task failure. Subsequent studies adopt this same formulation, inheriting the assumption that failure attribution can be uniquely determined \cite{zhang2025agentracer, kong2025aegis}.

\begin{figure*}[t] 
    \centering
    \includegraphics[width=.8\linewidth]{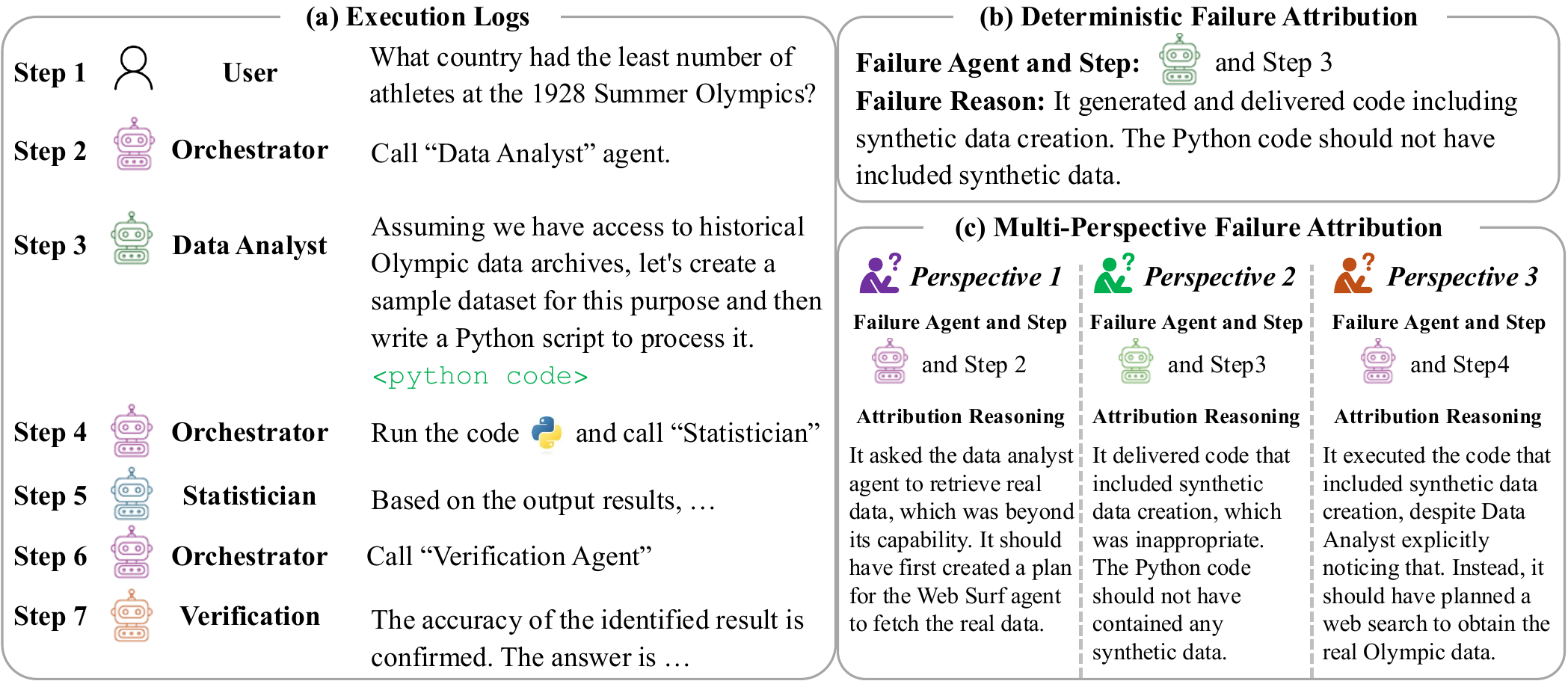}
    \vspace{-2ex}
    \caption{Motivating examples illustrating the multi-perspective nature of MAS failure attribution. (a) presents a simplified execution log of a MAS.
(b) presents the example of deterministic failure attribution existing approaches assume
(c) presents the example of multi-perspective failure attribution.}
    \label{fig:motivation}
    \vspace{-2ex}
\end{figure*}

However, we should rethink: \textit{Is the failure-inducing step for a MAS failure deterministically decided?} Consider the example in \Cref{fig:motivation}(a), where agents collaborate to address the user query. The Who\&When dataset and subsequent works implicitly assume that the oracle of failure-inducing steps is deterministic, thereby only one failure step are decided during annotation process (\Cref{fig:motivation}(b)).

In practice, the failure-inducing step of a MAS failure is often \textbf{not deterministic and admits multiple plausible attributions}. As illustrated in \Cref{fig:motivation}(c), different analysts may reasonably attribute the failure to different steps. 
This non-deterministic and multi-perspective nature arises from {the intrinsic interdependencies of MAS.} Multiple specialized agents can be interdependently composed into diverse, valid execution trajectories that aim toward the same goal. Consequently, failure attribution depends heavily on an observer’s internal reasoning of what constitutes a “correct” trajectory. In \Cref{fig:motivation}(c), the perspective difference emerges because one analyst expects a web-first strategy, while another considers a code-first approach acceptable. As further demonstrated in \Cref{sec:data-analysis}, even expert annotators exhibits perspective differences on the failure-inducing step in up to about 60\% of cases.

However, existing works largely overlook the inherently multi-perspective nature of MAS failure attribution, limiting their practical applicability. Since current benchmarks are built upon the assumption that the the oracle of a failure-inducing step is deterministic, they often misjudge reasonable predictions as incorrect, undermining evaluation reliability. Furthermore, models built on the assumption inherit biases that fail to capture diverse causal patterns, hindering effective debugging and refinement of MAS.

To address these limitations, we propose a more practical paradigm: \textit{\textbf{multi-perspective failure attribution}} for MAS. Under this paradigm, a failure attribution system identifies failure-inducing steps from multiple perspectives, each accompanied by explicit internal reasoning that explains the system’s judgment from that viewpoint. In contrast to prior work that frames failure attribution as a rigid, deterministic labeling task, our paradigm yields the multi-perspective attributions with concrete reasoning and actionable signals that more effectively support systematic MAS failure diagnosis and debugging.

This new paradigm necessitates a corresponding benchmark dataset and evaluation protocol. To this end, we introduce a new benchmark, \underline{\textbf{m}}ulti-\underline{\textbf{p}}erspective failure attribution \underline{\textbf{bench}} (\proposed), consisting of execution logs collected from diverse tasks and MAS configurations. Each log is annotated by multiple expert annotators who are rigorously screened via technical interviews and preliminary quality assessments on held-out test cases. The annotations include the failure-inducing step, and explicit internal reasoning that justifies the attribution from each annotator’s perspective. This expert-driven annotation process provides substantially stronger reliability than prior benchmarks, which largely rely on automated pipelines. We further design an evaluation protocol to assess whether current LLMs can perform multi-perspective failure attribution on \proposed. The evaluation focuses on two core capabilities: (1) the ability to identify failure-inducing steps from multiple perspectives, and (2) the ability to provide reliable and well-grounded reasoning to justify these attributions.

Through extensive experiments with state-of-the-art LLMs, we show that when sampled multiple times, LLMs effectively identify diverse failure attributions that correspond to different perspectives. Under our evaluation protocol, these attributions and the associated reasoning align closely with human expert's internal reasoning. Notably, these findings contrast sharply with prior results reported in deterministic benchmarks such as Who\&When, where LLM-based step-level failure attribution was shown to perform near random. Together, our results demonstrate that the limitations of LLMs in failure attribution observed in existing works stem not from insufficient model capability, but from biased evaluation assumptions that overlook the inherently multi-perspective nature of MAS failures. Moreover, we provide practical guidance for real-world MAS failure attribution: leveraging diverse sampling and complementary perspectives across models can more effectively capture diverse causal failure patterns and support systematic system debugging. The key contributions of this work are as follows:
\begin{itemize}[leftmargin=0.5cm]
    \item We identify a fundamental assumption in prior work that limits the practicality of MAS failure attribution and propose a more realistic task formulation, \textit{multi-perspective failure attribution}.
    \item We introduce a new benchmark and evaluation protocol specifically designed for multi-perspective failure attribution in MAS.
    \item We show that the prevailing belief that current LLMs perform poorly at failure attribution arises from flawed deterministic assumptions in existing benchmarks, rather than from inherent limitations of the models themselves.
    \item We provide practical guidance for real-world MAS failure attribution, demonstrating how leveraging diverse sampling strategies and complementary model perspectives can lead to more effective system diagnosis and debugging.
\end{itemize}

\section{Related Works}

\subsection{LLM-based Multi-Agent Systems}

Multi-agent systems built on LLMs have recently emerged as an effective paradigm for solving complex tasks through collaboration among multiple specialized agents \cite{hong2023metagpt, wu2024autogen, zhang2024aflow, song2024adaptive, ye2025mas}. While such systems benefit from modularity and role specialization, their intricate inter-agent dependencies often introduce reliability issues, where errors propagate across agents and failures arise from complex interactions \cite{cemri2025multi, zhangagent}. These challenges become more severe in long-horizon, multi-step reasoning settings, highlighting the need for systematic analysis of where and why MAS fail. Accordingly, this work focuses on failure attribution in MAS.

\subsection{Benchmarks for Failure Attribution of MAS}

Automatic failure attribution in MAS identifies failure-inducing steps from execution traces to support MAS debugging and refinement; accurately evaluating such systems is therefore essential. While several benchmarks have recently attracted attention, they remain limited in practicality for benchmarking real-world failure attribution systems. For example, MAST \cite{cemri2025multi} is restricted to execution-level failure mode analysis and thus cannot evaluate step-level failure attribution. TracerTraj \cite{zhang2025agentracer} and Aegis \cite{kong2025aegis} pursue finer-grained step- or agent-level attribution; however, their benchmarks are constructed through fully automated pipelines, raising concerns about annotation reliability given the complexity of MAS logs and causal failure patterns. Who\&When \cite{zhangagent} addresses these limitations by employing expert annotators to construct the benchmark dataset. Nevertheless, these benchmarks adopt a single-perspective formulation that assumes a deterministic oracle failure-inducing step, an assumption that conflicts with real-world MAS failures where multiple plausible attributions often coexist depending on an observer's perspective. Moreover, they frame failure attribution as a rigid labeling task and do not support evaluating the reasoning behind an attribution, which is essential for systematic failure diagnosis and debugging.

\begin{table}[h]
\centering
\vspace{-1ex}
\caption{Comparison between our benchmark and others.}
\vspace{-2ex}
\resizebox{1\linewidth}{!}{
\begin{tabular}{c|cccccc}
\toprule
& \makecell{Attribution \\ Granularity}   & \makecell{Annotation \\ Method}  & \makecell{Attribution \\ Reasoning Eval} & Perspective & Size \\
\midrule
\textbf{MAST \cite{cemri2025multi}} & Execution &  Hybrid &  \ding{55} &  Single  & 1,600 \\
\textbf{Who\&When \cite{zhangagent}} & Step  &  Expert &  \ding{55} & Single &  184 \\
\textbf{TracerTraj \cite{zhang2025agentracer}} & Step & Automated & \ding{55} & Single & 2,500\\
\textbf{Aegis \cite{kong2025aegis}} & Agent & Automated  &  \ding{55} & Single & 9,500 \\
\midrule
\textbf{\proposed} & Step & Expert & \ding{51} & Multi   & 289 \\
\bottomrule
\end{tabular}}
\vspace{-1ex}
\label{tab:comparison}
\end{table}

To address these limitations, we introduce the first multi-perspective failure attribution benchmark for MAS, constructed through a rigorous expert annotator recruitment and labeling process to ensure consistently high-quality annotations. We further propose a new evaluation protocol tailored to the multi-perspective setting, enabling systematic comparison of failure attribution systems under attribution ambiguity. In addition, our benchmark provides annotator rationales for each attribution, along with an evaluation framework that explicitly assesses both attribution quality and underlying reasoning. It enables structured and fine-grained analysis of failure patterns across diverse agents, tasks, and perspectives. A detailed comparison between existing benchmarks and our contributions is summarized in \Cref{tab:comparison}.

\subsection{Annotator Disagreement}

Recent literature has increasingly adopted a perspectivist approach that treats annotator disagreement as a meaningful signal rather than noise to be resolved \cite{fleisig2024perspectivist, wan2025noise}. Prior work has explored various modeling strategies to preserve and leverage diverse viewpoints \cite{dehghan2025dealing, xu2025modeling, homayounirad2025will}. However, these studies have largely focused on socially subjective tasks—such as hate speech or radical content detection \cite{dehghan2025dealing, riabi2025beyond, orlikowski2025beyond, fleisig2023majority}—where judgments are often grounded in personal, cultural, or normative values rather than technical reasoning.

In contrast, we present the first study to investigate annotator disagreement in the context of failure attribution for MAS executions, a setting that requires complex technical reasoning and system diagnosis. Moreover, we introduce a novel benchmark that explicitly captures the multi-perspective nature of MAS failures. By doing so, we extend the scope of perspectivism beyond social subjectivity to address ambiguity arising from the interpretability and diagnostic complexity of real-world MAS.

\section{Task Formulation}

\subsection{Preliminaries}

\noindent \textbf{Multi-Agent Systems.} \@
We adopt a widely used turn-based LLM multi-agent protocol \cite{hong2023metagpt, li2023camel, wu2024autogen, zhangagent}. Specifically, we consider an LLM-powered MAS $\mathcal{M}$ consisting of a set of $N$ agents, denoted by $\mathcal{N} = {1, 2, \ldots, N}$, that operate in discrete time steps. At each time step, exactly one agent takes an action according to a turn-based interaction rule. The system is formally defined as $\mathcal{M} = \left\langle \mathcal{N}, S, A, P, \phi \right\rangle$, where $S$ denotes the set of environment states, $A$ is the global action space, and each agent $i \in \mathcal{N}$ is associated with a subset of allowable actions $A_i \subseteq A$. The function $\phi(t)$ specifies the active agent at time step $t$. The state transition dynamics are governed by $P(s_{t+1} \mid s_t, a_t, \phi(t))$, which defines the probability of transitioning to state $s_{t+1}$ given the observable state $s_t$ and the action $a_t$ taken by the active agent at time $t$.
An execution of the system over a task is represented as a finite trajectory
\begin{equation}
\small
\tau = {(s_1, a_1, \phi(1)), \ldots, (s_T, a_T, \phi(T))},
\end{equation}
where $T$ denotes the total number of steps.

\noindent \textbf{Deterministic Failure Attribution.}
Existing failure attribution benchmarks and methods assume that, for each failed execution $\tau$, there exists a deterministic set of failure-inducing steps that can be uniquely identified. Under this assumption, failure attribution is formulated as a mapping $f_{\text{det}}(\tau) \rightarrow \mathcal{F}^\ast$, where $\mathcal{F}^\ast \subseteq \{1, \ldots, T\}$ denotes a fixed set of time steps whose associated agent actions $(a_t, \phi(t))$ are deemed responsible for the failure.
This formulation treats failure attribution as an objective labeling problem with a single correct attribution set.
In practice, most benchmarks further simplify this formulation by assuming $|\mathcal{F}^\ast| = 1$, selecting a single step as failure-inducing \cite{zhangagent, zhang2025agentracer}.

\subsection{Multi-Perspective Failure Attribution}

\noindent \textbf{Definition.} \@
In contrast, we argue that MAS failure attribution is inherently non-deterministic. 
Due to the existence of multiple valid execution trajectories and diverse expectations of a correct strategy, a failed execution may admit multiple plausible failure attributions, each grounded in a distinct analytical perspective.

We define \textit{multi-perspective failure attribution} as the task of identifying a set of plausible failure-inducing steps for a given failed execution $\tau$, together with their associated rationales. 
Formally, the output is a set $\mathcal{F}(\tau) = \{(t_k, r_k)\}_{k=1}^{K}$,
where each $t_k \in \{1, \ldots, T\}$ denotes a failure-inducing time step, and $r_k$ is an explicit rationale explaining why the agent action $(a_{t_k}, \phi(t_k))$ is considered a failure-inducing step under a particular perspective $k$. 
The cardinality $K$ is not fixed and may vary across executions.
Importantly, multiple attributions in $\mathcal{F}(\tau)$ can be simultaneously valid, even when they identify different steps—and thus different agents via $\phi(t_k)$—as responsible for the failure.

\noindent \textbf{Task Objective.} \@
Given a failed execution trace $\tau$, the objective of multi-perspective failure attribution is to predict a set of plausible failure attributions $\widehat{\mathcal{F}}(\tau)$ that aligns with human expert judgments. Unlike deterministic settings, success is not defined by matching a single oracle label, but by the model’s ability to recover multiple reasonable attributions and their associated rationales.

This formulation explicitly acknowledges the multi-perspective nature of MAS failure attribution and reframes the task from identifying a single failure-inducing step to characterizing a space of plausible causal explanations.

\section{\proposed: Multi-Perspective Failure Attribution Benchmark}

\subsection{Benchmark Dataset Construction}

\begin{figure*}[t] 
    \centering
    \includegraphics[width=.7\linewidth]{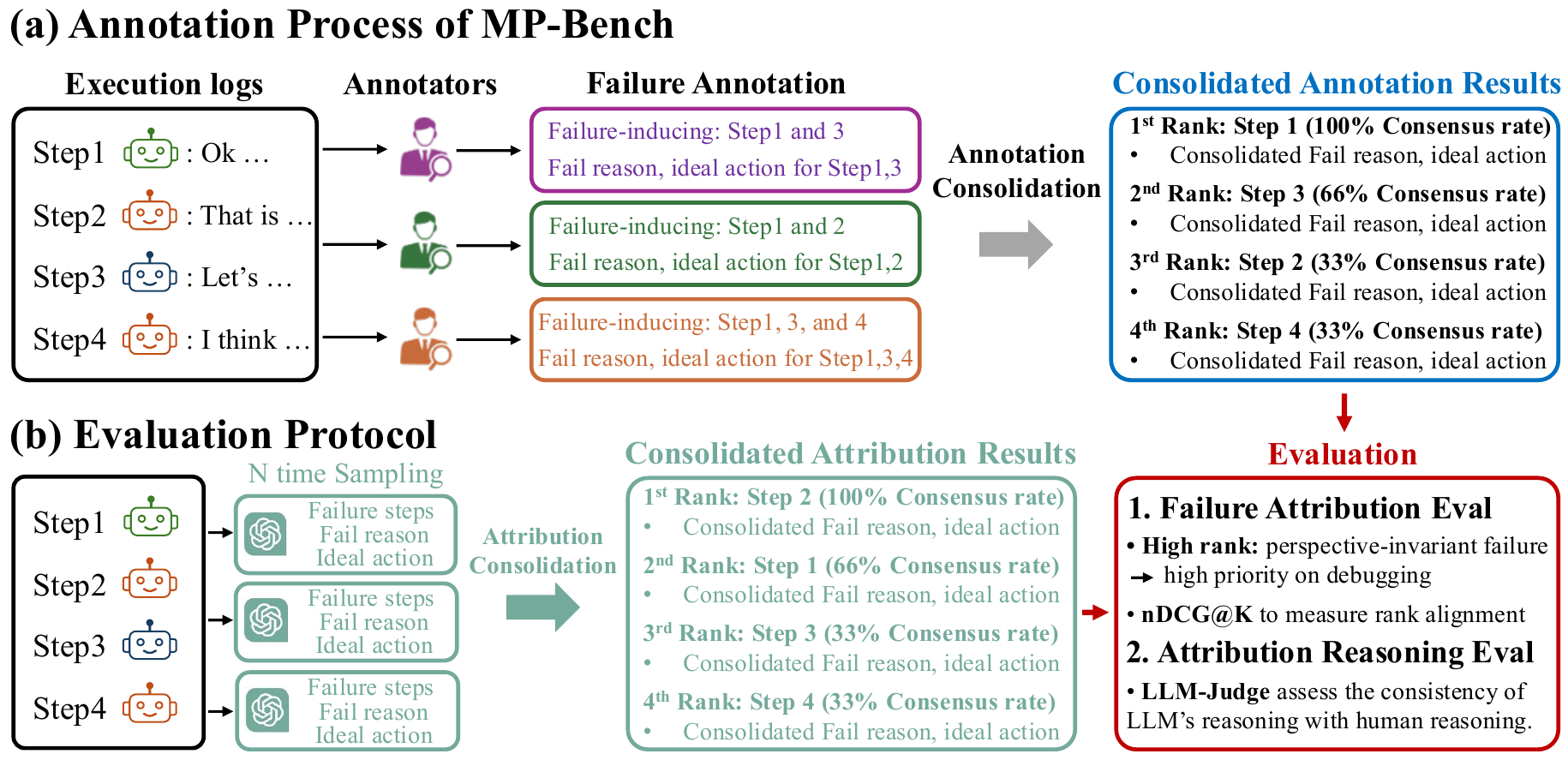}
    \vspace{-2ex}
    \caption{Overall framework of the (a) annotation process and (b) evaluation protocol of \proposed.}
    \vspace{-2ex}
    \label{fig:overall}
\end{figure*}

This section describes the construction of \proposed, a benchmark dataset designed for \textit{multi-perspective failure attribution} in MAS. Our dataset is built to capture diverse failure patterns across systems, tasks, and annotator perspectives, while providing structured rationales for the step-level agent actions. The overall procedure is summarized in \Cref{fig:overall}.

\subsubsection{Multi-Agent Systems and Execution Log Collection}
To ensure diversity in system design and execution behavior, following \cite{zhangagent}, we include both hand-crafted and fully-automated MAS configurations. Specifically, we collect 289 logs from a total of 121 distinct MAS configurations and various tasks. For hand-crafted MAS, we use MAgenticOne \cite{fourney2024magentic}. For fully-automated MAS, we use CaptainAgent \cite{song2024adaptive}. For tasks, we use GAIA \cite{mialon2023gaia} and AssistantBench \cite{yoran2024assistantbench}, which are widely utilized to evaluate the agentic capabilities of LLM-based MAS. 

As the first benchmark to address multi-perspective failure attribution, we prioritize the quality of annotation over dataset scale. We argue that for evaluating complex diagnostic capabilities such as failure attribution—where understanding nuanced causal patterns and diverse interpretations is critical—a carefully curated set of expert annotations provides more reliable evaluation signals than automated labels that may lack grounding or consistency.

While our dataset of 289 execution logs is smaller than recent automated benchmarks (e.g., TracerTraj with 2,500 instances and Aegis with 9,500), these works rely on fully automated data generation pipelines that cannot ensure annotation reliability for the complex reasoning patterns inherent in MAS failures. In contrast, we employ a rigorous expert annotation process with multiple annotators per instance, capturing diverse perspectives and explicit reasoning for each attribution. Compared to Who\&When, the closest prior work that also uses manual curation, our benchmark provides 57\% more instances (289 vs. 184) while additionally introducing the multi-perspective paradigm. This design enables broader coverage of failure patterns and more realistic evaluation of attribution systems in practice.

\subsubsection{Expert Annotator Recruitment}
We recruit three expert annotators on Upwork\footnote{https://www.upwork.com/} platform. To ensure high-quality and technically grounded annotations, we prioritize candidates with professional experience in LLMs, AI systems, or agent-based frameworks. We adopt a rigorous two-stage screening process.
In the first stage, we conduct technical interviews to assess candidates’ conceptual understanding of the task. Applicants are asked to explain, in their own words, what MAS failure attribution entails, and to describe how they would approach annotating a turn-based MAS execution log. In the second stage, we assess annotation quality using test cases. Specifically, we provide several MAS execution logs and ask candidates to annotate them according to a detailed predefined task description (see \Cref{fig:annotation-description} in Appendix). The authors manually review the submitted annotations, focusing on correctness, consistency, and the clarity of reasoning. Only candidates who pass both stages are selected, resulting in three verified annotators. 

\subsubsection{Annotation Task} 
For each execution log, annotators are provided with the complete MAS 
execution trace and task description. Each trace consists of a sequence 
of agent actions following a turn-based protocol. Annotators independently 
annotate every step, assigning the following for each:

\begin{itemize}[leftmargin=0.5cm]
\item \textbf{Failure-inducing Step:} a binary label indicating whether the step is failure-inducing.
\item \textbf{Failure Reason:} a free-text explanation justifying the failure annotation.
\item \textbf{Ideal Action:} a free-text description of the action that the agent should ideally have taken.
\end{itemize}

Critically, we provide no potential bias or guiding perspectives to annotators, allowing them to develop independent interpretations based on their expertise. This design genuinely captures the multi-perspective nature of MAS failure attribution, as annotators may reasonably disagree on which steps are failure-inducing based on different assumptions about correct execution strategies. 
The \textit{failure reason} and \textit{ideal action} fields explicitly capture each annotator's internal reasoning, grounding why a particular agent action is deemed problematic from their perspective. This enables \proposed~to evaluate failure attribution systems on two dimensions: 
(1) the ability to identify failure-inducing steps across multiple perspectives, and (2) the ability to generate well-grounded reasoning that aligns with expert judgment. This evaluation framework moves beyond existing benchmarks that treat failure attribution as deterministic classification, instead assessing both attribution accuracy and reasoning quality.

\subsubsection{Annotation Consolidation}
\label{sec:consolidation}

After collecting annotations from all annotators, we consolidate them into a ranking-based format. For each execution log, we compute a \textit{consensus rate} for every step, defined as the fraction of annotators who labeled it as failure-inducing. Steps are then ranked by consensus rate, creating an actionable prioritization for debugging.

Steps with 100\% consensus represent perspective-invariant failure that is consistently recognized across all annotators. Such steps represent the most salient and unambiguous failure points, and therefore should be assigned the highest priority during MAS refinement. In contrast, steps labeled as failures by only a subset of annotators reflect perspective-dependent failure modes. These steps may be considered failures under certain interpretations of a correct execution trajectory, but not under others, and thus typically warrant lower refinement priority. Importantly, only steps that are labeled as failures by at least one annotator are included in the ranking; steps that are never identified as failures are excluded. By ranking failure-inducing steps according to their consensus rates, our formulation transforms multi-perspective annotations into an actionable prioritization signal for developers, aligning failure attribution with practical MAS debugging workflows.

While binary failure annotations naturally aggregate into consensus-based rankings, failure reasons and ideal actions are free-text and require different treatment. We consolidate these textual explanations using LLM-assisted summarization. Specifically, we prompt an LLM to synthesize a unified failure reason and ideal action for each failure-inducing step, conditioned on all annotator-provided explanations. The LLM is instructed to comprehensively cover the diverse rationales without introducing new information. We primarily use GPT-5.1~\cite{openai2025gpt51} for consolidation. To verify robustness to this choice, we additionally evaluate consolidations generated by GPT-4.1~\cite{openai2025gpt41} and Claude-Sonnet-4.5~\cite{claude2025sonnet45}. We observe that model rankings and performance patterns remain consistent across all three consolidation methods, confirming that our results are not sensitive to the choice of consolidation model. Detailed results are presented in Appendix~\ref{sec:sensitivity-llm-choice}.

\vspace{-2ex}
\subsection{Annotation Quality and Effort}

\noindent \textbf{Dataset Composition.}
\proposed~consists of 169 hand-crafted MAS executions (average 33 interaction steps) and 120 automatically generated executions (average 8 steps), totaling 289 instances with triple expert annotation.

\noindent \textbf{Annotator Qualifications.}
All annotators possess substantial technical expertise in AI and multi-agent systems: Annotator 1 (7 years in AI engineering and NLP), Annotator 2 (10+ years in applied AI and LLM-based 
agentic systems), and Annotator 3 (5 years in AI/ML development, LLMs, and AI agents). This ensures annotations reflect genuine expert judgment on MAS failure patterns.

\noindent \textbf{Annotation Effort.}
Each hand-crafted execution requires approximately 31 minutes per annotator on average, while automatic executions requires approximately 14 minutes, reflecting the complexity difference in execution length and interaction patterns. Across all 289 instances and three annotators, 
the total annotation effort amounts to approximately 346 expert-hours (115 hours per annotator). Annotators were compensated at competitive rates commensurate with their expertise levels.

This substantial time investment prioritizes annotation quality and multi-perspective coverage over dataset scale, ensuring the reasoning quality essential for reliable evaluation of failure attribution systems. While this approach limits immediate scalability, it establishes a 
gold-standard benchmark for validating future automated annotation methods.


\vspace{-2ex}
\subsection{Data Analysis}
\label{sec:data-analysis}

\begin{figure}[h] 
    \centering    \includegraphics[width=.9\linewidth]{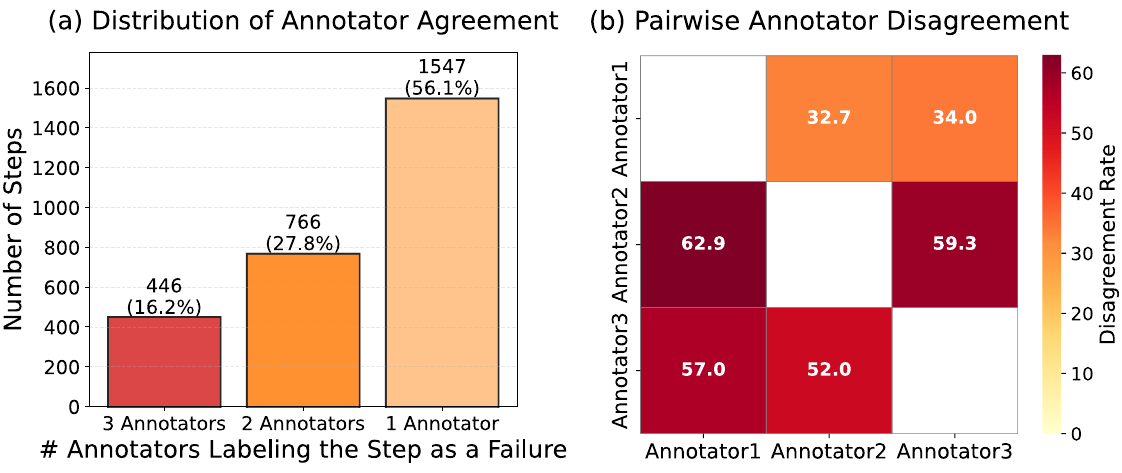}
    \vspace{-2ex}
    \caption{Analysis of \proposed~highlighting the multi-perspective nature of MAS failure attribution. (a) Distribution of steps grouped by annotator consensus on failure annotations. (b) Inter-annotator disagreement rates for failure annotations.}
    \label{fig:disagree}
\end{figure}

We analyze the annotation statistics of \proposed, focusing on the degree of consensus among annotators regarding failure step annotations. As shown in \Cref{fig:disagree}(a), only 16.2\% of steps are annotated as failure-inducing by all three annotators. In contrast, 27.8\% of steps are labeled as failures by exactly two annotators, while a substantial 56.1\% are labeled as failures by only a single annotator. This distribution indicates that failure attribution is highly dependent on annotator perspective, underscoring the limitations of deterministic assumptions adopted by existing benchmarks. Similarly, \Cref{fig:disagree}(b) reports inter-annotator disagreement rates, defined as cases where one annotator marks a step as failure while another does not. The results show that the disagreement can reach up to 60\% across annotators, further confirming the multi-perspective nature of MAS failure attribution.

\vspace{-2ex}
\section{Evaluation Protocol}

This section describes the evaluation protocol for benchmarking MAS failure attribution systems on \proposed. The protocol is explicitly designed to address the following questions:

\begin{itemize}
\item Do LLM-based failure attribution systems capture failures from multiple perspectives? (\Cref{sec:failure-attribution-evaluation})
\item Do LLMs provide reliable reasoning for failure attribution? (\Cref{sec:attribution-grounding-evaluation})
\end{itemize}

\noindent An overview of the evaluation protocol is shown in \Cref{fig:overall}(b).

\subsection{Failure Attribution Evaluation}
\label{sec:failure-attribution-evaluation}

Existing failure attribution benchmarks formulate evaluation as a classification task, assessing whether a system correctly identifies a single failure-inducing step. Such deterministic evaluation is insufficient in our setting, as a single execution may admit multiple plausible failure attributions depending on perspective.

To address the first question, in \proposed, we adopt a ranking-based evaluation that measures the consistency between the rankings produced by LLM-based failure attribution systems and the ground-truth rankings derived from human annotations. Specifically, we run each LLM $N$ times with a sampling temperature of $\tau$, generating multiple independent failure attribution outputs. Each output consists of predicted failure-inducing steps along with their associated failure reasons, and ideal actions. Following the same consolidation procedure used for human annotations, we aggregate these outputs into a single predicted ranking based on consensus rates.

To compare rankings produced by annotators and LLMs, we employ normalized Discounted Cumulative Gain (nDCG@K). This metric evaluates the agreement between the predicted ranking and the ground-truth ranking, with greater emphasis on correctly prioritizing highly salient failure steps. As a result, nDCG@K aligns evaluation with practical MAS debugging workflows, where addressing the most critical failures first is essential.

\subsection{Attribution Reasoning Evaluation}
\label{sec:attribution-grounding-evaluation}

In the multi-perspective failure attribution task, beyond identifying failure-inducing steps, LLMs generate explicit attribution reasoning including failure reasons and ideal actions. Evaluating this reasoning quality is essential for assessing whether LLMs can serve as reliable 
failure attribution systems.

Following the same procedure as annotation consolidation (Section~\ref{sec:consolidation}), we consolidate the attribution reasoning generated from $N$ LLM runs. We then adopt an LLM-as-a-Judge framework using GPT-5.1 as the judge model. For each failure-inducing step, the judge assesses consolidated predictions against consolidated ground truths based on: (1) consistency with human judgments, (2) grounding in execution context, (3) explanatory adequacy, and (4) reasonableness of proposed actions. The judge assigns scores (1-10) for failure reason, ideal action, and overall reasoning quality. We calculate the average score of overall reasoning quality across all steps and executions. Detailed judge prompts are provided in \Cref{fig:llm-as-a-judge-attribution-reasoning-eval} in Appendix.

To verify robustness to judge model choice, we additionally evaluate using GPT-4.1 and Claude-Sonnet-4.5 as alternative judges. We find that model rankings remain highly consistent across all three judges, confirming that our evaluation results are not sensitive to this choice. Detailed results are presented in Appendix~\ref{sec:sensitivity-llm-choice}.

\begin{table*}[t]
\centering
\caption{Failure attribution and attribution reasoning performance comparison.}
\label{tab:main-results}
\resizebox{0.8\textwidth}{!}{
\begin{tabular}{lcccccccccc}
\toprule
& \multicolumn{5}{c}{Hand-Crafted MAS} & \multicolumn{5}{c}{Automatic MAS} \\
\cmidrule(lr){2-6} \cmidrule(lr){7-11}

& \multicolumn{4}{c}{Failure Attribution Eval} & \multicolumn{1}{c}{Attr. Reason. Eval}
& 
\multicolumn{4}{c}{Failure Attribution Eval} & \multicolumn{1}{c}{Attr. Reason. Eval} \\

\cmidrule(lr){2-5}  \cmidrule(lr){7-10} 

& \multicolumn{2}{c}{nDCG@5 (↑)} & \multicolumn{2}{c}{nDCG@full (↑)}
& LLM-Judge 
& \multicolumn{2}{c}{nDCG@5 (↑)} & \multicolumn{2}{c}{nDCG@full (↑)}
& LLM-Judge \\

\cmidrule(lr){2-3} \cmidrule(lr){4-5}
\cmidrule(lr){7-8} \cmidrule(lr){9-10}

Model
& Exp & Linear & Exp & Linear
& Avg. Score (↑)
& Exp & Linear & Exp & Linear
& Avg. Score (↑) \\

\midrule
Random            & 0.1275 & 0.2368 & 0.1476 & 0.2228 & {-}  & 0.3147 & 0.4153 & 0.3207 & 0.415  & {-} \\
\midrule
GPT-4.1           & 0.4313 & 0.5386 & 0.4717 & 0.4961 & 7.63                                 & 0.6755 & 0.695  & 0.6761 & 0.6851 & 7.78                  \\
GPT-5.1           & 0.3747 & 0.5164 & 0.4526 & 0.553  & 7.62                  & 0.7844 & \textbf{0.8134} & 0.7856 & \textbf{0.8051} & \textbf{8.25}                   \\
O3-mini           & 0.4367 & 0.5299 & 0.4487 & 0.4436 & 7.48                  & 0.3944 & 0.4085 & 0.3936 & 0.3979 & 7.37                      \\
Claude-Sonnet-4.5 & \textbf{0.4397} & \textbf{0.5814} & \textbf{0.506}  & \textbf{0.571}  & \textbf{7.95}                               & \textbf{0.7894} & 0.7998 & \textbf{0.792}  & 0.7936 & 8.23                                  \\
\midrule
Qwen3-8B          & 0.2944 & 0.3945 & 0.338  & 0.3885 & 6.11                                  & 0.6681 & 0.6739 & 0.6809 & 0.6784 & 6.54                  \\
GPT-oss-120B      & 0.4245 & 0.5456 & 0.4852 & 0.5405 & 7.51                                  & 0.703  & 0.7173 & 0.7134 & 0.7166 & 7.92                               \\

\bottomrule
\end{tabular}
}
\end{table*}

\vspace{-2ex}
\section{Experiment}

\subsection{Experimental Setup}

\subsubsection{Failure Attribution Systems Details}

We implement failure attribution systems using a diverse set of LLMs, including both proprietary and open-source models: GPT-4.1 \cite{openai2025gpt41}, GPT-5.1 \cite{openai2025gpt51}, o3-mini \cite{openai2025o3mini}, GPT-oss-120B \cite{agarwal2025gpt}, Claude-Sonnet-4.5 \cite{claude2025sonnet45}, and Qwen3-8B \cite{yang2025qwen3}. For failure attribution, we adopt the All-at-Once method \cite{zhangagent} in a zero-shot setting, which provides a favorable balance between efficiency and accuracy. Unless otherwise specified, we set the number of LLM runs to $N=3$, the sampling temperature to $\tau=1.0$, and the maximum output length to 4,196 tokens. Proprietary models are accessed via their official APIs \cite{openai-api, claude-api}. For the open-source GPT-oss-120B, we use the Together.ai API \cite{togetherai-api}. Qwen3-8B is run locally on our own machine equipped with an NVIDIA RTX A6000 GPU.

\subsubsection{Evaluation Details}

For failure attribution evaluation, we report nDCG under multiple settings, including nDCG@5 and nDCG@full, with both exponential and linear gain variants. This design accounts for the fact that exponential gain places greater emphasis on steps with high consensus rates. By additionally reporting full-ranking and linear-gain metrics, we provide a more comprehensive evaluation that also reflects performance on lower-consensus, perspective-dependent failure steps. For attribution reasoning evaluation, we use GPT-5.1 as the consolidation and judge model with $\tau = 0.0$ to ensure stable and deterministic judgments. Based on manual inspection, we find that the judge model produces reliable evaluation results.

\subsection{Experimental Results}

\noindent \textbf{Main Results.} \@ From the evaluation on \proposed, we derive the following key observations.

\begin{figure}[h] 
    \centering    \includegraphics[width=.9\linewidth]{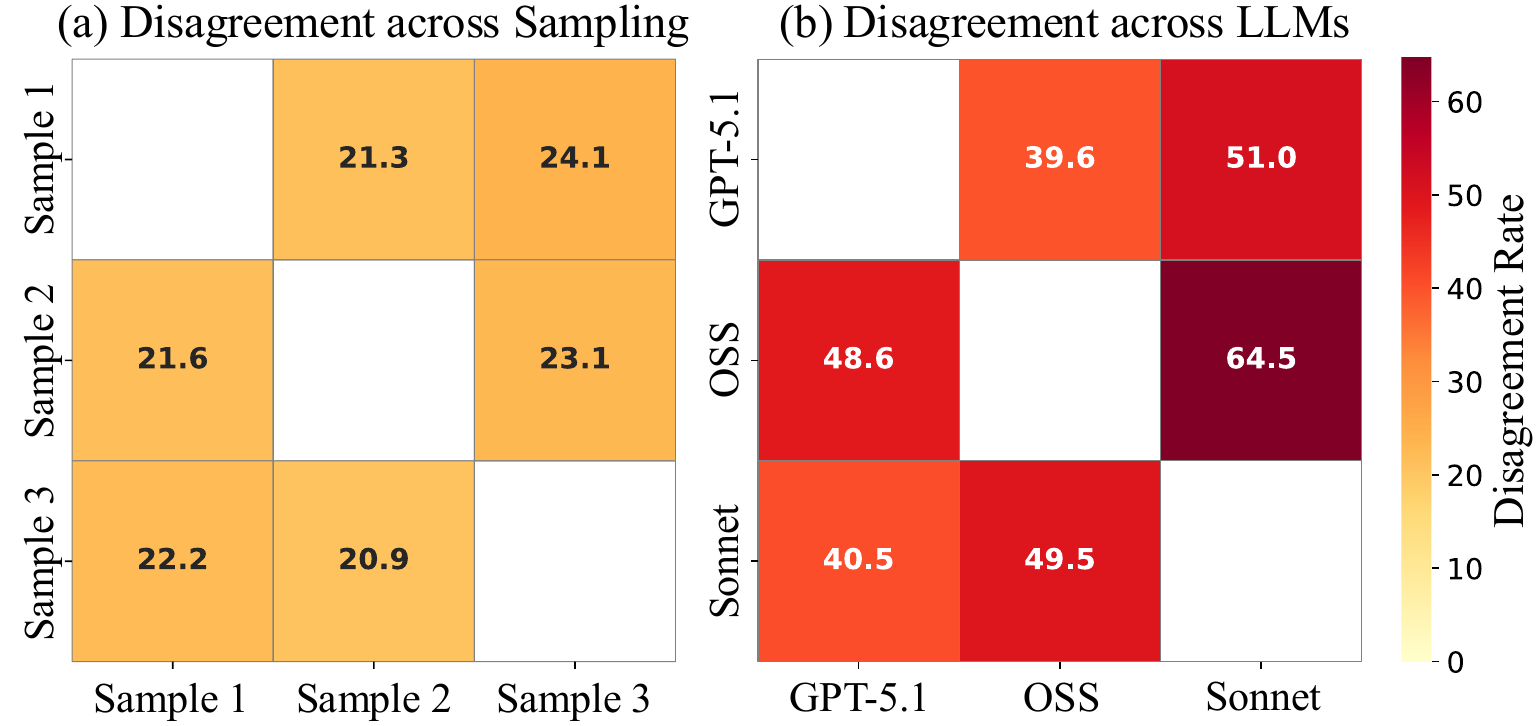}
    \caption{Disagreement analysis of the LLM-based failure attribution system. (a) Pairwise disagreement rates across different samplings of GPT-5.1. (b) Pairwise disagreement rates across different LLMs. }
    \label{fig:disagree-llm}
\end{figure}

\textbf{First, stochastic sampling of LLMs naturally reflects diverse perspectives in failure attribution.}
When sampling multiple failure attributions from LLMs, we observe substantial disagreement among the predicted failure-inducing steps (e.g., GPT-5.1 in \Cref{fig:disagree-llm}(a)), a pattern that closely mirrors the disagreement observed among human expert annotators. This alignment provides further evidence that failure attribution in MAS is inherently non-deterministic and admits multiple plausible perspectives, rather than a single deterministic root cause.

\textbf{Second, under a multi-perspective evaluation setting, LLMs perform failure attribution effectively when sampled multiple times with high temperature.}
As shown in \Cref{tab:main-results}, when LLMs are sampled multiple times ($N=3$) with a high sampling temperature ($\tau=1$), they successfully identify multiple reasonable failure-inducing steps across different perspectives. Among the evaluated models, Claude-Sonnet-4.5 consistently achieves the strongest performance, whereas Qwen3-8B exhibits limited failure attribution and reasoning capabilities, likely due to its relatively small model size. Notably, we observe that stronger reasoning-oriented or more recent models do not necessarily guarantee superior performance in this setting. That said, the resulting nDCG scores from all models are substantially higher than those of a random baseline. These results stand in contrast to prior findings from deterministic benchmarks such as Who\&When \cite{zhangagent}, which suggested that LLMs struggle with step-level failure attribution by exhibiting near-random performance.

\textbf{Third, LLM-generated failure attributions are accompanied by reliable and well-grounded reasoning.}
In \Cref{tab:main-results}, we observe high LLM-as-a-Judge scores when comparing the attribution reasoning produced by LLMs with that of human annotators. This indicates that LLM-generated multi-perspective attributions are not only plausible in terms of identified failure steps, but are also supported by reasoning that aligns well with human expert judgments. Consequently, these attributions can serve as meaningful diagnostic signals for MAS developers during system debugging. 

\textbf{Overall, our results challenge the prevailing assumption that MAS failure attribution is deterministic and that in-context LLMs are inherently ineffective at this task.}
Instead, we show that MAS failure attribution exhibits an inherently multi-perspective nature, and that LLMs possess sufficient capability to capture this structure when evaluated under an appropriate benchmark and evaluation protocol. The limitations reported in prior work therefore stem not from deficiencies in model capability, but from biased evaluation assumptions that overlook the multi-perspective nature of MAS failures.

\subsubsection{Sensitivity Analyses on Sampling Temperature}

\begin{table}[h]
\centering
\small 
\caption{Performance comparison (nDCG@5 with exponential gain) across different sampling temperatures $\tau$. Bold values indicate the best performance within each model.}
\label{tab:sensitivity-temp}

\resizebox{0.8\columnwidth}{!}{ 
\begin{tabular}{lcccc|c}
\toprule
& \multicolumn{5}{c}{\textbf{Total (Hand-Crafted + Automatic)}} \\
\cmidrule(lr){2-6}
 & \textbf{$\tau=1$} & \textbf{$\tau=0.7$} & \textbf{$\tau=0.5$} & \textbf{$\tau=0.3$} & \textbf{$\tau=0$}  \\
\midrule
GPT-4.1           & 0\textbf{.5349} & 0.5227 & 0.5314 & 0.5219 & 0.5162 \\
GPT-5.1           & 0.552  & 0.562  & 0.5561 & \textbf{0.5628} & 0.552  \\
Claude-Sonnet-4.5 & \textbf{0.5916} & 0.5863 & 0.588  & 0.5894 & 0.5874 \\
\midrule
Qwen3-8B          & 0.4553 & 0.4619 & 0.4692 & \textbf{0.4803} & 0.3886 \\
GPT-oss-120B      & 0.5449 & 0.5439 & 0.5458 & \textbf{0.5554} & 0.5432 \\
\bottomrule
\end{tabular}
} 
\end{table}

We analyze the effect of sampling temperature ($\tau$) on multi-perspective failure attribution performance by varying $\tau$ from 0 to 1 and evaluating nDCG@5 with exponential gain across different LLMs. We exclude the o3-mini model from this analysis, as it only supports $\tau = 1$. As shown in \Cref{tab:sensitivity-temp}, performance at $\tau = 0$ is consistently worse than at higher temperatures. This suggests that fully deterministic decoding constrains the model’s output space, thereby limiting the diversity of failure attributions. In contrast, once stochasticity is introduced ($\tau \geq 0.3$), performance improves substantially. However, we do not observe a monotonic performance increase as $\tau$ continues to rise from 0.3 to 1.0. This indicates that while avoiding overly restrictive decoding is important for capturing diverse attribution perspectives, increasing the temperature beyond a moderate level offers limited additional benefits; in practice, moderately non-deterministic sampling is sufficient.

\subsubsection{Sensitivity Analyses on Number of LLM Sampling}
\label{sec:sensitivy-number-of-llm-salmping}

\begin{table}[h]
\centering
\small 
\caption{Performance comparison (nDCG@5 with exponential gain) across different number of LLM runs $N$. Bold values indicate the best performance within each model.}
\label{tab:sensitivity-num-llm}

\resizebox{\columnwidth}{!}{ 
\begin{tabular}{lcccccc}
\toprule
& \multicolumn{3}{c}{\textbf{Hand-Crafted}} & \multicolumn{3}{c}{\textbf{Automatic}} \\
\cmidrule(lr){2-4} \cmidrule(lr){5-7}
 & \textbf{$N=3$} & \textbf{$N=5$} & \textbf{$N=10$} & \textbf{$N=3$} & \textbf{$N=5$} & \textbf{$N=10$} \\
\midrule
GPT-4.1           & 0.4359 & 0.4384 & \textbf{0.4622} & 0.6874 & 0.6945 & \textbf{0.7201} \\
GPT-5.1           & 0.3806 & 0.3993 & \textbf{0.4198} & \textbf{0.7819} & 0.7801 & 0.7769 \\
O3-mini           & 0.4323 & 0.4725 & \textbf{0.4862} & 0.4125 & 0.4254 & \textbf{0.4604} \\
Claude-Sonnet-4.5 & 0.4472 & 0.4672 & \textbf{0.4753} & 0.766  & 0.7761 & \textbf{0.7861} \\
\midrule
Qwen3-8B          & 0.2951 & 0.3241 & \textbf{0.3442} & 0.662  & 0.69   & \textbf{0.7227} \\
GPT-oss-120B      & 0.4224 & 0.4574 & \textbf{0.4779} & 0.7115 & 0.7193 & \textbf{0.7269} \\
\bottomrule
\end{tabular}
} 
\end{table}

We further examine the impact of the number of LLM runs ($N$) on multi-perspective failure attribution performance. We vary $N$ from 3 to 10 and report nDCG@5 with exponential gain for different LLMs. As shown in \Cref{tab:sensitivity-num-llm}, increasing $N$ leads to consistent performance improvements across models. Notably, performance continues to improve as $N$ increases, indicating that aggregating a larger number of samples yields more stable and reliable failure attribution rankings. This trend suggests that multiple runs help mitigate variability arising from limited sampling and enable more robust aggregation of perspective-dependent failure signals.

\subsubsection{Analyses on Multi-LLM Failure Attribution} 
\label{sec:analyses-multi-llm}

\begin{table}[h]
\centering
\caption{Performance comparison between single-LLM and multi-LLM failure attribution systems. For brevity, we denote GPT-4.1, GPT-5.1, GPT-OSS-120B, and Claude-Sonnet-4.5 as G-4.1, G-5.1, OSS, and Sonnet, respectively. }
\label{tab:multi-llm}
\vspace{-2ex}
\resizebox{0.8\columnwidth}{!}{ 
    \begin{tabular}{cccccc}
    \toprule
    \multicolumn{4}{c}{\textbf{Model Combination}} & \multicolumn{2}{c}{\textbf{nDCG@5 (Exp)}} \\
    \cmidrule(lr){1-4} \cmidrule(lr){5-6}
    \textbf{G-4.1} & \textbf{G-5.1} & \textbf{OSS} & \textbf{Sonnet} & \textbf{Hand-Crafted} & \textbf{Automatic} \\
    \midrule
    \cmark & \xmark & \xmark & \xmark & 0.4313 & 0.6755 \\
    \xmark & \cmark & \xmark & \xmark & 0.3747 & 0.7844 \\
    \xmark & \xmark & \cmark & \xmark & 0.4245 & 0.703 \\
    \xmark & \xmark & \xmark & \cmark & 0.4397 & 0.7894 \\
    \midrule
    \cmark & \cmark & \cmark & \xmark & 0.479 & 0.8 \\
    \cmark & \cmark & \xmark & \cmark & 0.4909 & 0.8142 \\
    \cmark & \xmark & \cmark & \cmark & 0.\textbf{496} & 0.7937 \\
    \xmark & \cmark & \cmark & \cmark & 0.4906 & \textbf{0.8228} \\
    \bottomrule
    \end{tabular}
} 
\end{table}

We study whether collaborating multiple LLMs improves failure attribution performance, motivated by the hypothesis that different models encode distinct reasoning patterns and thus offer complementary attribution perspectives. To evaluate this, we compare single-LLM systems with multi-LLM systems constructed from GPT-4.1, GPT-5.1, GPT-OSS-120B, and Claude-Sonnet-4.5. We focus on combinations that achieve strong performance in \Cref{tab:main-results}. In the single-LLM setting, each model is sampled three times ($N=3$), whereas in the multi-LLM setting, each model is sampled once. All results are evaluated using nDCG@5 with exponential gain.

The results in \Cref{tab:multi-llm} show that multi-LLM systems consistently outperform their single-LLM counterparts. Importantly, the largest performance gains are obtained when combining models from different families. For instance, mixtures of Claude, GPT-based, and OSS models achieve the highest attribution performance, suggesting that cross-family diversity is particularly effective for failure attribution.

This observation is further supported by the disagreement analysis in \Cref{fig:disagree-llm}. We find that pairwise disagreement across multiple samplings of the same LLM is substantially lower than the disagreement observed across different LLMs. This indicates that repeated sampling primarily captures stochastic variation, whereas combining distinct models introduces qualitatively different attribution perspectives. Taken together, these findings suggest that the performance gains of multi-LLM systems stem from complementary internal reasoning patterns across models, rather than from increased sampling alone.

\subsection{Practical Guidance for Practitioners}

Based on our experimental findings, we provide practical guidance for designing effective and reliable automatic failure attribution systems for MAS.

\textbf{First, MAS failure attribution is inherently multi-perspective.} Rather than relying on a single deterministic attribution result, practitioners should emphasize aggregating attributions from multiple perspectives, which enables capturing diverse causal patterns and supports more effective MAS debugging and refinement.

\textbf{Second, sampling temperature should be set above zero.} Fully deterministic decoding can overly constrain the model’s output space, limiting the diversity of failure attributions produced by LLMs.

\textbf{Third, combining LLMs from different model families is strongly recommended.} Models from different families tend to provide complementary perspectives, and aggregating their attributions leads to more informative and robust failure diagnosis.

\textbf{Finally, when higher reliability is required, increasing the number of samples is beneficial.} Aggregating a larger set of attribution results enables more stable and accurate prioritization of failure-inducing steps.

\section{Conclusion}

We revisit failure attribution in MAS and show that deterministic formulations fail to capture the inherently multi-perspective nature of real-world MAS failures. To address this limitation, we propose multi-perspective failure attribution and introduced \proposed, the first benchmark and evaluation framework designed for this setting. \proposed~combines expert annotations and multi-perspective evaluation to enable faithful assessment of failure attribution quality and reasoning. Our experiments reveal that prior claims about LLMs’ limitations in failure attribution largely stem from restrictive benchmark assumptions, and that modern LLMs can produce diverse, human-aligned attributions when evaluated appropriately. Overall, our work underscores the need to rethink failure attribution benchmarks and evaluation practices, and provides a practical foundation for reliable MAS diagnosis and debugging.

\section{Limitations and Future Work}

While \proposed~establishes the first rigorous benchmark for multi-perspective failure attribution, it also has several limitations that suggest directions for future work.

\noindent \textbf{Domain Coverage.} \@  
Our benchmark currently focuses on general-purpose assistant tasks. Extending the benchmark to more specialized domains, such as scientific research, software engineering, and creative tasks, would broaden its applicability and enable a more comprehensive evaluation of failure attribution methods.

\noindent \textbf{Scale and Framework Diversity.} \@  
Although our quality-first approach ensures reliable and high-fidelity annotations, it limits the scale of the benchmark and the diversity of covered multi-agent system frameworks. Future work could investigate hybrid expert–automated annotation pipelines to improve scalability while maintaining annotation quality, as well as incorporate a wider range of MAS frameworks.

Despite these limitations, \proposed~provides a strong gold-standard foundation for multi-perspective failure attribution research and serves as a valuable resource for advancing this line of work.

\bibliography{refer.bib}

\appendix
\newpage
\clearpage

\section{Use of Generative AI Tools}

Generative AI tools are used solely for minor editing purposes during the manuscript preparation, such as improving sentence fluency and correcting typographical errors. They are not used to generate new content, analyses, or experimental results.

\section{Ethics Statement}

This dataset is constructed exclusively using publicly available tasks and publicly accessible multi-agent systems. It does not contain any personally identifiable information. To the best of our knowledge, the dataset does not raise any ethical concerns.

\section{Additional Experimental Results}

\subsection{Sensitivity Analyses on LLM choice for Consolidation and LLM-as-a-Judge}
\label{sec:sensitivity-llm-choice}

In our main experiments, we use GPT-5.1~\cite{openai2025gpt51} for 
both annotation consolidation (Section~\ref{sec:consolidation}) and 
LLM-as-a-Judge evaluation (Section~\ref{sec:attribution-grounding-evaluation}). 
To verify that our evaluation is robust to these choices, we conduct 
comprehensive sensitivity analyses using GPT-4.1~\cite{openai2025gpt41} 
and Claude-Sonnet-4.5~\cite{claude2025sonnet45} as alternative models. 
With three consolidation models and three judge models, we evaluate 
all nine possible combinations.

\noindent \textbf{Absolute Score Variation across Judges.} \@
Tables~\ref{tab:sensi-1} and \ref{tab:sensi-2} 
present attribution reasoning scores across all nine configurations for 
hand-crafted and automatic MAS executions, respectively. We observe that 
GPT-4.1 as judge consistently assigns higher scores than GPT-5.1 and 
Claude-Sonnet-4.5. Given that GPT-5.1 and Claude-Sonnet-4.5 possess 
stronger reasoning capabilities, they likely evaluate more rigorously, 
resulting in more conservative scoring. Nonetheless, scores across all 
configurations consistently exceed 7.0, indicating high-quality 
attribution reasoning from all evaluated systems regardless of judge 
choice.

\noindent \textbf{Ranking Consistency: Hand-Crafted MAS.} \@
Despite variation in absolute scores, model rankings remain highly 
consistent. For hand-crafted MAS executions, we compute Kendall's 
rank correlation coefficient across all $\binom{9}{2} = 36$ 
configuration pairs. The results demonstrate strong consistency:
\begin{itemize}[leftmargin=0.5cm]
    \item Average $\tau = 0.896$
    \item Minimum $\tau = 0.733$, Maximum $\tau = 1.00$
    \item 31 of 36 pairs achieve statistical significance ($p < 0.05$)
    \item All 36 pairs achieve $p < 0.1$
\end{itemize}
These correlations substantially exceed conventional thresholds for 
strong agreement, confirming that model rankings are robust to 
evaluation configuration.

\noindent \textbf{Ranking Consistency: Automatic MAS.} \@
For automatic MAS executions, we observe similarly strong consistency:
\begin{itemize}[leftmargin=0.5cm]
    \item Average $\tau = 0.904$
    \item Minimum $\tau = 0.733$, Maximum $\tau = 1.00$
    \item 30 of 36 pairs achieve $p < 0.05$
    \item All 36 pairs achieve $p < 0.1$
\end{itemize}

\noindent \textbf{Conclusion.} \@
While absolute scores vary with judge model choice—reflecting 
differences in evaluation stringency—relative rankings remain 
highly stable across all configurations. This consistency demonstrates 
that our conclusions about model performance are not artifacts of 
specific consolidation or judge choices, but reflect genuine 
differences in failure attribution quality. The strong agreement 
(average $\tau \simeq 0.90$ for both settings) provides confidence that 
practitioners can reliably identify top-performing systems regardless 
of evaluation setup.

\begin{table*}[t]
\centering
\caption{Performance comparison across consolidation and judge model 
combinations for hand-crafted MAS executions. Sonnet denotes 
Claude-Sonnet-4.5. Values in parentheses indicate model rankings.}
\label{tab:sensi-1}

\resizebox{1\textwidth}{!}{\begin{tabular}{lccccccccc}
\toprule
 & \multicolumn{3}{c}{Consolidation: GPT-4.1} 
 & \multicolumn{3}{c}{Consolidation: GPT-5.1} 
 & \multicolumn{3}{c}{Consolidation: Sonnet} \\
\cmidrule(lr){2-4} \cmidrule(lr){5-7} \cmidrule(lr){8-10}

& Judge: GPT-4.1 & Judge: GPT-5.1 & Judge: Sonnet
& Judge: GPT-4.1 & Judge: GPT-5.1 & Judge: Sonnet
& Judge: GPT-4.1 & Judge: GPT-5.1 & Judge: Sonnet \\
\midrule
GPT-4.1              & 8.49 (3) & 7.52 (3) & 7.07 (2) & 8.60 (3) & 7.63 (2) & 7.20 (2) & 8.48 (3) & 7.44 (3) & 7.14 (2) \\
GPT-5.1              & 8.57 (2) & 7.53 (2) & 7.04 (3) & 8.68 (2) & 7.62 (3) & 7.14 (4) & 8.56 (2) & 7.49 (2) & 7.03 (3) \\
O3-mini              & 8.22 (5) & 7.31 (5) & 6.93 (5) & 8.32 (5) & 7.48 (5) & 7.06 (5) & 8.13 (5) & 7.23 (5) & 6.92 (5) \\
Claude-Sonnet-4.5  & 8.70 (1) & 7.82 (1) & 7.27 (1) & 8.83 (1) & 7.95 (1) & 7.50 (1) & 8.64 (1) & 7.71 (1) & 7.32 (1) \\
\midrule
Qwen3-8B               & 7.45 (6) & 5.97 (6) & 5.82 (6) & 7.54 (6) & 6.11 (6) & 5.96 (6) & 7.37 (6) & 5.92 (6) & 5.77 (6) \\
GPT-oss-120B
                              & 8.37 (4) & 7.40 (4) & 7.03 (4) & 8.46 (4) & 7.51 (4) & 7.18 (3) & 8.27 (4) & 7.34 (4) & 6.97 (4) \\
\bottomrule
\end{tabular}}
\end{table*}

\begin{table*}[t]
\centering
\caption{Performance comparison across consolidation and judge model 
combinations for automatic MAS executions. Sonnet denotes 
Claude-Sonnet-4.5. Values in parentheses indicate model rankings.}
\label{tab:sensi-2}
\resizebox{1\textwidth}{!}{\begin{tabular}{lccccccccc}
\toprule
 & \multicolumn{3}{c}{Consolidation: GPT-4.1} 
 & \multicolumn{3}{c}{Consolidation: GPT-5.1} 
 & \multicolumn{3}{c}{Consolidation: Sonnet} \\
\cmidrule(lr){2-4} \cmidrule(lr){5-7} \cmidrule(lr){8-10}

& Judge: GPT-4.1 & Judge: GPT-5.1 & Judge: Sonnet
& Judge: GPT-4.1 & Judge: GPT-5.1 & Judge: Sonnet
& Judge: GPT-4.1 & Judge: GPT-5.1 & Judge: Sonnet \\
\midrule
GPT-4.1           & 8.5 (4)  & 7.66 (4) & 7.15 (4) & 8.69 (3) & 7.78 (4) & 7.38 (4) & 8.43 (4) & 7.49 (4) & 7.13 (4) \\
GPT-5.1           & 8.84 (1) & 8.06 (1) & 7.45 (2) & 9.01 (2) & 8.25 (1) & 7.68 (2) & 8.81 (1) & 8.03 (1) & 7.48 (1) \\
O3-mini           & 8.13 (5) & 7.21 (5) & 6.86 (5) & 8.28 (5) & 7.37 (5) & 7.06 (5) & 8.03 (5) & 7.16 (5) & 6.75 (5) \\
Claude-Sonnet-4.5 & 8.8 (2)  & 8.01 (2) & 7.5 (1)  & 9.02 (1) & 8.23 (2) & 7.73 (1) & 8.77 (2) & 7.94 (2) & 7.47 (2) \\
\midrule
Qwen3-8B          & 7.54 (6) & 6.38 (6) & 6.03 (6) & 7.74 (6) & 6.54 (6) & 6.24 (6) & 7.43 (6) & 6.17 (6) & 5.93 (6) \\
GPT-OSS-120B      & 8.57 (3) & 7.87 (3) & 7.25 (3) & 8.66 (4) & 7.92 (3) & 7.46 (3) & 8.5 (3)  & 7.66 (3) & 7.25 (3) \\
\bottomrule
\end{tabular}}
\end{table*}

\subsection{Scaling $N$ in Multi-LLM Failure Attribution}

\begin{figure}[h] 
    \centering    \includegraphics[width=.85\linewidth]{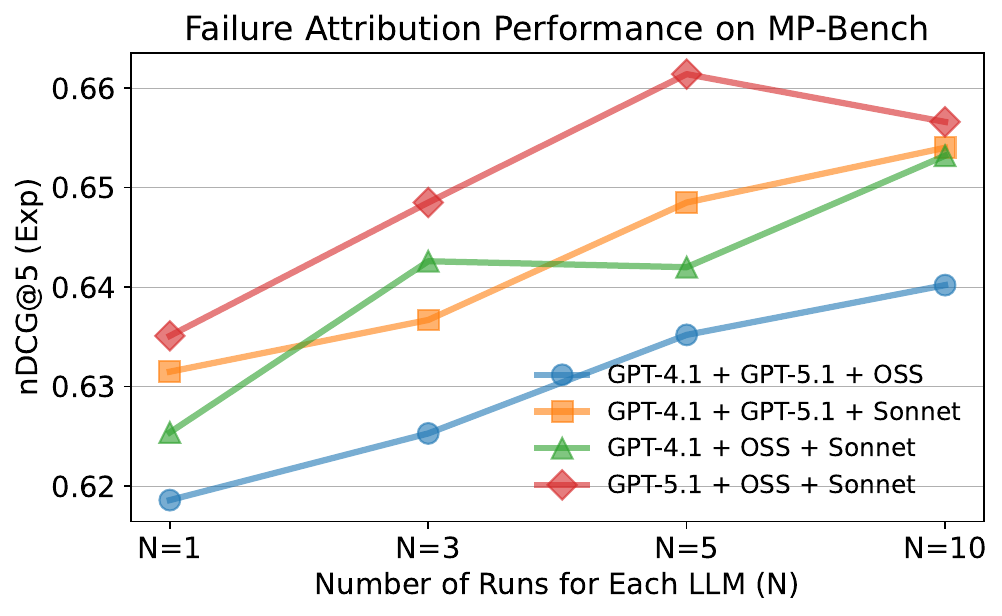}
    \caption{Failure attribution performance across varying numbers of LLM runs ($N$) on \proposed. OSS denotes GPT-OSS-120B, and Sonnet denotes the Claude-Sonnet-4.5 model.}
    \label{fig:multillm}
\end{figure}

We investigate whether the scaling behavior with respect to $N$ observed in \Cref{sec:sensitivy-number-of-llm-salmping} also holds in the multi-LLM failure attribution setting. For multi-LLM configurations, we follow the same setup as in \Cref{sec:analyses-multi-llm}, where three LLMs collaboratively perform failure attribution. We vary $N$ from 1 to 10, corresponding to a total of 3 to 30 LLM runs (3 $\times$ $N$). Performance is evaluated using the average nDCG@5 with exponential gain across both hand-crafted and automatically generated MAS datasets. 
As shown in \Cref{fig:multillm}, increasing the number of samples ($N$) leads to steady performance improvements, consistent with prior observations.

\begin{figure*}[t] 
    \centering
    \includegraphics[width=.7\linewidth]{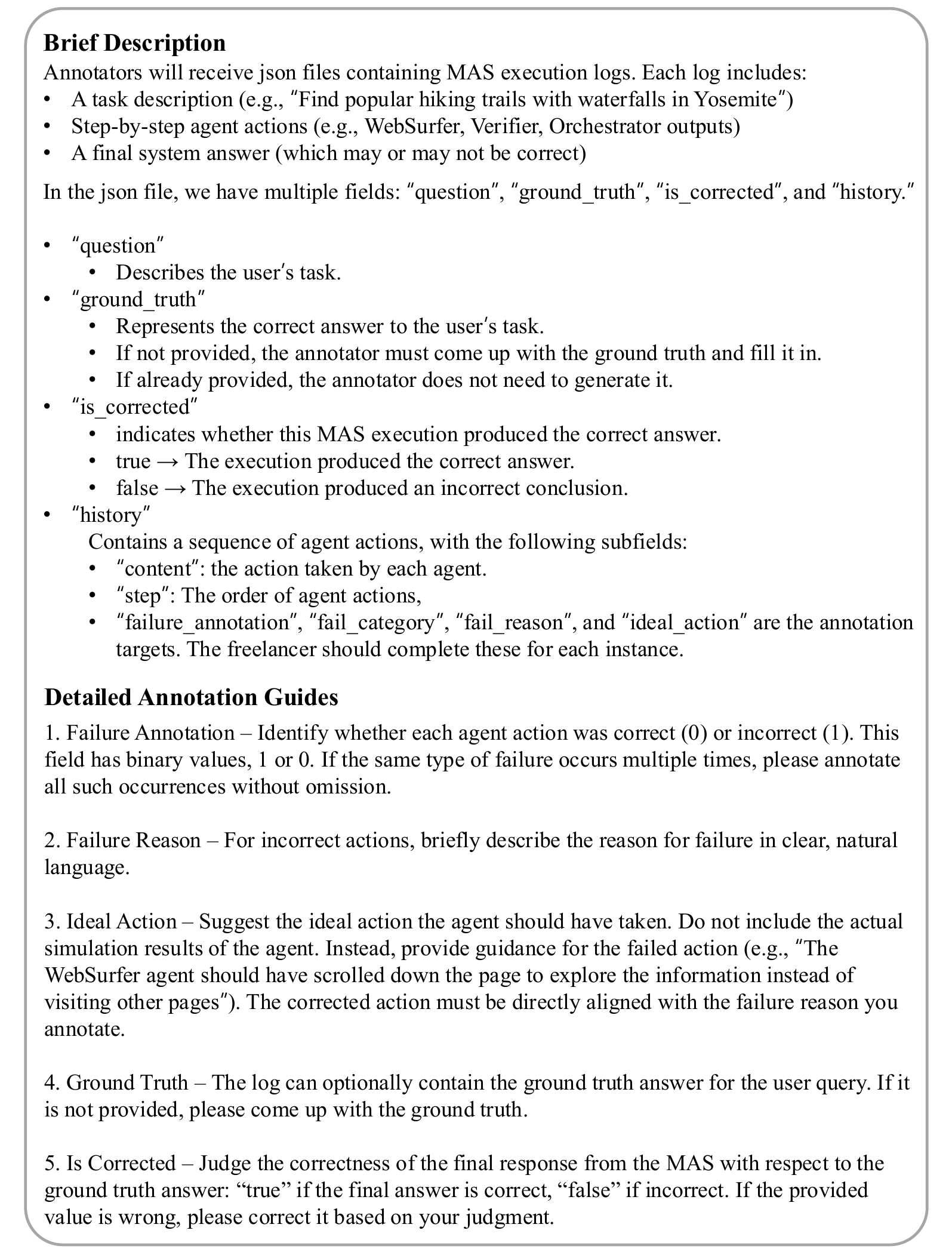}
    \vspace{-2ex}
    \caption{Annotation description provided to annotators.}
    \label{fig:annotation-description}
\end{figure*}

\begin{figure*}[p]
    \centering
    \begin{tcolorbox}[
        colback=white,       
        colframe=gray!80,    
        arc=3mm,             
        boxrule=1pt,         
        width=\textwidth,    
        left=5pt, right=5pt, top=5pt, bottom=5pt 
    ]
        You are an AI assistant tasked with analyzing a multi-agent conversation history when solving a real-world problem. The problem is: \textcolor{blue}{\texttt{\{problem\}}}.

        Identify all agents who made errors, at which steps each error occurred, and analyze the nature of these errors.
        
        Here's the conversation: \\
        -------------------------- \\
        \textcolor{blue}{\texttt{\{chat history content\}}} \\
        -------------------------- \\
        Based on this conversation, please provide a detailed analysis of the mistakes in the following JSON format. The output must be a list of objects, ordered by the step number. Each object must include:
        \\
        1. agent\_name: The name of the agents who made any mistakes during the conversation. Directly output the name of the Expert.\\
        2. step\_number: The integer step number where the mistake occurred (e.g., 0, 1, 2...). For example, in a conversation structured as follows: \\
        \{ \\  
            "agent a": "xx", \\
            "agent b": "xxxx", \\ 
            "agent c": "xxxxx", \\
            "agent a": "xxxxxxx" \\ 
        \}, \\ 
        
        the textual chat history is structured as follows: \\
        
        Step 0: agent a: x  \\
        Step 1: agent b: xxxx \\
        Step 2: agent c: xxxxx \\ 
        Step 3: agent a: xxxxxxx \\

        each entry represents a 'step' where an agent provides input. The 'x' symbolizes the speech of each agent. Please determine the step numbers where the mistakes occurred.
        
        Please specify the step number as presented."
        
        3. failure\_reason: Briefly describe the reason for failure in clear, natural language.
        
        4. ideal\_action: Suggest the ideal action the agent should have taken. Do not include actual simulation results; instead, provide guidance aligned with the failure reason (e.g., 'The WebSurfer agent should have scrolled down the page to explore the information instead of visiting other pages').
        
        
        

        Please answer ONLY in the following JSON format:

        [\\
            \{ \\
                "agent\_name": "AgentName1", \\
                "step\_number": "only integer", \\
                "failure\_reason": "Description of why it failed", \\
                "ideal\_action": "Guidance on what should have been done", \\
            \}, \\
            ... \\
        ] \\
        Please order the results by the step number. Do not include any other text or comments.

    \end{tcolorbox}
    \caption{Prompt for failure attribution method (All At Once \cite{zhangagent}).}
    \label{fig:prompt_all_at_once}
\end{figure*}

\begin{figure*}[p]
    \centering
    \begin{tcolorbox}[
        colback=white,       
        colframe=gray!80,    
        arc=3mm,             
        fontupper=\small, 
        boxrule=1pt,         
        width=\textwidth,    
        left=5pt, right=5pt, top=5pt, bottom=5pt 
    ]

You are an expert evaluator assessing the quality of a large language model's failure attribution reasoning by comparing it with a human expert annotator's annotation.

Your goal is NOT to check for exact wording matches, but to evaluate whether the model provides a plausible, well-grounded reasoning that aligns with the human annotator's judgment.

---

Task: \textcolor{blue}{\texttt{\{question\}}}

Failure Step: \textcolor{blue}{\texttt{\{step\_num\}}}

Execution Log (context around step \textcolor{blue}{\texttt{\{step\_num\}}}):

\textcolor{blue}{\texttt{\{chat history content\}}}

---

Model Prediction: \\
- Failure Reason: \textcolor{blue}{\texttt{\{pred\_fail\_reason\}}} \\
- Ideal Action: \textcolor{blue}{\texttt{\{pred\_ideal\_action\}}} \\

Human Expert Annotation: \\
- Failure Reason: \textcolor{blue}{\texttt{\{{GT\_fail\_reason}\}}} \\
- Ideal Action: \textcolor{blue}{\texttt{\{{GT\_ideal\_action}\}}} \\

---

Evaluation Criteria:

Evaluate the model's prediction by jointly considering the failure reason and ideal action, based on the following aspects:

1. Reasoning Alignment\\
   Does the model capture the core reasoning behind why the step is considered a failure and what should have been done instead, as reflected in the human annotation? \\

2. Faithfulness to Execution Context\\   
   Is the model's reasoning grounded in the actual execution context of this step?\\
   - Does it rely only on information available in the execution trace?\\
   - Does it avoid introducing unsupported assumptions or hindsight knowledge?\\

3. Coverage and Completeness  \\
   Does the model address the key issues and considerations raised by the annotator, even if expressed differently?\\

4. Plausibility of Ideal Action\\
   Is the proposed ideal action reasonable, actionable, and consistent with the annotator’s intent?\\

---

Scoring Guidelines (1–10 scale):

- 9–10 (Excellent):  \\
  Strong alignment with the annotator’s reasoning; well-grounded, faithful to the execution context, and captures the key rationale and ideal action clearly.\\

- 7–8 (Good):  \\
  Largely aligned with minor omissions or differences; reasoning is plausible and context-faithful.\\

- 5–6 (Moderate):\\
  Partially aligned but missing important aspects or containing some questionable assumptions.\\

- 3–4 (Poor):\\
  Weak alignment; reasoning diverges significantly from the annotator or lacks grounding in the execution context. \\

- 1–2 (Very Poor): \\
  Largely incorrect, unfaithful to the execution context, or fails to provide a meaningful explanation. \\

---

Only return the JSON object. Do not include any additional text. Output Format (JSON only):\\
\{
    "overall\_score": "integer between 1 and 10", \\
    "fail\_reason\_score": "integer between 1 and 10",\\
    "ideal\_action\_score": "integer between 1 and 10", \\
    "reasoning": "Concise explanation of the judgment" \}
        
    \end{tcolorbox}
    \vspace{-3ex}
    \caption{LLM-as-a-Judge prompt for attribution reasoning evaluation.}
    \label{fig:llm-as-a-judge-attribution-reasoning-eval}
\end{figure*}

\bibliographystyle{ACM-Reference-Format}
\citestyle{acmauthoryear}

\end{document}